\documentclass[oneside]{article}

\usepackage{PRIMEarxiv}

\usepackage{times}
\usepackage{url}
\usepackage[hidelinks]{hyperref}
\usepackage[utf8]{inputenc}
\usepackage[small]{caption}
\usepackage{graphicx}
\usepackage{amsmath}\usepackage{amsfonts}
\usepackage{amsthm}
\usepackage{booktabs}
\usepackage{algorithm}
\usepackage{algorithmic}

\usepackage{makecell}
\usepackage{multirow,tabularx}
\usepackage{array}
\newcolumntype{L}[1]{>{\raggedright\let\newline\\\arraybackslash\hspace{0pt}}m{#1}}
\newcolumntype{C}[1]{>{\centering\let\newline\\\arraybackslash\hspace{0pt}}m{#1}}
\newcolumntype{R}[1]{>{\raggedleft\let\newline\\\arraybackslash\hspace{0pt}}m{#1}}

\usepackage{subfigure}

\title{Explainable Graph Neural Networks for Observation Impact Analysis in Atmospheric State Estimation}
\author {
    Hyeon-Ju Jeon\textsuperscript{\rm 1,}\thanks{Corresponding author: Hyeon-Ju Jeon (Tel.: +82-2-6480-6425)}
    Jeon-Ho Kang\textsuperscript{\rm 1}, 
    In-Hyuk Kwon\textsuperscript{\rm 1}, 
    O-Joun Lee\textsuperscript{\rm 2}
}

\author{
Hyeon-Ju Jeon${}^{\dagger}$, Jeon-Ho Kang, In-Hyuk Kwon\\
Data Assimilation Group\\
Korea Institute of Atmospheric Prediction Systems\\
Seoul, Republic of Korea\\
\texttt{\{hjjeon,jhkang,ihkwon\}@kiaps.org}
\And
O-Joun Lee\\
Dept. of Artificial Intelligence\\
The Catholic University of Korea\\
Bucheon, Republic of Korea\\
\texttt{ojlee@catholic.ac.kr}
}

\begin{document}
\maketitle

\begin{abstract}
This paper investigates the impact of observations on atmospheric state estimation in weather forecasting systems using graph neural networks (GNNs) and explainability methods. We integrate observation and Numerical Weather Prediction (NWP) points into a meteorological graph, extracting $k$-hop subgraphs centered on NWP points. Self-supervised GNNs are employed to estimate the atmospheric state by aggregating data within these $k$-hop radii. The study applies gradient-based explainability methods to quantify the significance of different observations in the estimation process. Evaluated with data from 11 satellite and land-based observations, the results highlight the effectiveness of visualizing the importance of observation types, enhancing the understanding and optimization of observational data in weather forecasting.

\let\thefootnote\relax
\footnotetext{${}^{\dagger}$ Correspondence: \texttt{hjjeon@kiaps.org}; Tel.: +82-2-6480-6497} 
\end{abstract}


\section{Introduction}
Weather forecasting, a critical component in industries like transportation and manufacturing, relies heavily on Numerical Weather Prediction (NWP) systems, which are based on 3D physical models and dynamical equations \cite{Stulec2019, Kotsuki2019}. 
For NWP systems to predict future atmospheric states effectively, they require accurate current atmospheric states as initial values. 
This necessity underscores the importance of a data assimilation (DA) system, which approximates the true atmospheric states by merging observations with prediction results from dynamical models \cite{Kwon2018}. The integration of a wide range of observations, from sources like aircraft, radiosondes, and satellites, is crucial for enhancing the DA system's accuracy \cite{Kang2018}.

Traditional methods to assess the impact of observations on weather forecasts include forecast sensitivity to observation (FSO) and its variations, such as ensemble FSO and hybrid FSO \cite{Kotsuki2019, Kalnay2012, Buehner2018}. 
These methods compute the gradient of the forecast with respect to the assimilated observations within the DA system but are limited by their dependency on the system's structural changes.

Our study proposes a novel approach using Graph Neural Networks (GNNs) \cite{Jeon2021, Hoang2023, Lee2021} to estimate the impact of observations independently of the system's structure. 
GNNs have been increasingly employed in meteorological predictions, including solar radiation and sea surface temperature predictions, by capturing variable interactions in neighboring regions \cite{Jeon2022, Ma2023, Yang2018}. 
The GraphCast \cite{Lam2023} model, for instance, transforms the NWP system's 3D grid into a hierarchical graph to capture long-range spatial interactions using GNNs. However, it does not incorporate the latest observations in its predictive model. 
To our knowledge, no existing model fuses observations with NWP grids in a graph format for current atmospheric state estimation using GNNs.

Additionally, we apply explainability methods to evaluate the impact of observations on current atmospheric state estimation \cite{Yuan2022}. These methods, including model gradient analysis and input perturbation, have been previously used for validating deep learning models in Earth system science \cite{Pope2019, Ying2019, Vu2020, Irvine2011}. 
Our study extends the use of these methods to feature analysis in atmospheric science, providing a novel perspective.

The contributions of this paper include:
\begin{itemize}
    \item Defining a meteorological graph that includes real atmospheric state and observational data, addressing the challenge of unstructured data.
    \item Developing a self-supervised graph convolutional network (GCN) model for atmospheric state estimation, demonstrating superior performance over other baseline models.
    \item Using explainability methods to estimate and visually represent the impact of observations on accurate atmospheric state predictions.
\end{itemize}

\section{Atmospheric State Estimation}

\subsection{Composing Meteorological Contexts}

Weather forecasting heavily relies on both local and global weather conditions, but estimation of the current atmospheric state is predominantly influenced by nearby observations. In the Korean Integrated Model (KIM), observations impact NWP grid points within a 50km radius. This study treats atmospheric state estimation as a node-level regression task, aiming to estimate the states of NWP grid points at time \( t \) using data from NWP grid points at time \( t-1 \) and observations at time \( t \).

A meteorological graph, \( \mathcal{N}=(V, E,\lambda_v) \), was constructed, comprising observation points and NWP grid points as nodes, where \( V \ni v_i \) represents the nodes, \( E \) the edges, and \( \lambda_v \) a function mapping nodes to their types. Adjacency between nodes is determined by a 50km radius proximity.

In this graph \( \mathcal{N} \), various weather variables are assigned as node attributes \( a_i \) to each node \( v_i \). The type and quantity of these variables differ based on node types (observation or NWP points) and observation types (e.g., IASI, GK2A, etc.). 
To standardize different feature vector sizes, a projection layer \( p(\mathcal{N}, v_i)_{\lambda_v} \) maps each node \( v_i \) to a fixed size vector \( h_i \in \mathbb{R}^d \), with \( d \) being the embedding dimension.

Given the extensive scale of the 3D NWP grid points, which are uniformly distributed across the globe at 50km intervals, the meteorological graph is large, posing challenges for GNNs application. 
In addition, local observations have more influence on the estimation of the atmospheric state than the broader global weather context.
Therefore, the study extracts ego-centric subgraphs centered on NWP points from the main graph. Each subgraph \( g_m(v_i) \) comprises \( k \)-hop neighbors of \( v_i \) and their interconnecting edges. Utilizing these subgraphs as individual samples, we effectively aggregate local weather contextual information within the \( k \)-hop range. This approach allows for the application of GNNs to atmospheric state estimation while managing computational costs. 

\subsection{Pre-training with Node Feature Reconstruction}


Observational data in meteorology, such as temperature, can have varying implications depending on the surrounding weather conditions. 
For instance, a temperature of 305K might signify different things in tropical regions compared to mid-latitude areas. 
To interpret these variations, GCNs are pre-trained on a node attribute reconstruction task. This process enables the determination of the specific meanings of observations in diverse meteorological contexts.

The GCN-based graph encoder takes a meteorological context subgraph with node initial feature vectors ($h_i$) passed through the projection layer and composes final vector representations of nodes ($H_i$). 
This pre-training phase involves training node representations in meteorological contexts alongside the attribute reconstruction task. 
The graph encoder, utilizing GCN layers, concurrently learns node features and graph structures. 
The graph encoder is formulated as:
\begin{align}
	H^{(l+1)} = \sigma(\tilde{D}^{-\frac{1}{2}} \tilde{A} \tilde{D}^{-\frac{1}{2}} H^{(l)}W^{(l)}),
\end{align} 
where $W^{(l)}$ indicates a weight matrix in $l^{th}$ layer, $H^{(l)}$ is the feature matrix generated by the $l^{th}$ layer.  
In addition, $\tilde{D}_{ii}=\sum_j\tilde{A}_{ij}$, and $\tilde{A}=A+I_N$ is the adjacency matrix of the context subgraph $g_m$, where $I_N$ indicates the identity matrix.
Also, $H^{(0)}$ is the set of $h_i$.

The objective of this pre-training is to approximate the reconstructed node attributes to the actual attributes as closely as possible, minimizing reconstruction error.
This error is quantified using the L2 loss, with the objective function defined as:
\begin{align}
	L_{ssl} = \lVert a, \hat{a}\rVert_2^2+\psi\lVert W\rVert_2^2,
\end{align} 
where $a$ and $\hat{a}$ are the actual and predicted node attributes, respectively, and $\psi$ is a weight for the regularization term.

\subsection{Estimating Current Atmospheric States}

In our approach to atmospheric state estimation, node representations from the graph encoder are transformed into representations of ego-centric subgraphs through graph pooling. Subsequently, a Multi-Layer Perceptron (MLP) is employed to estimate the current atmospheric states of the central nodes, which are NWP points, based on these subgraph representations. 
This process can be formulated as:
\begin{align}
	Z = \textrm{MLP}(\textrm{POOL}(H^{(n)})),
\end{align} 
where $H^{(n)} = \textrm{GCN}(h,A)$ is the final node representations from the GCN layers.
$\textrm{POOL}(\cdot)$ indicates the graph pooling layer, which conducts average pooling for node representations in context subgraphs.
Thereby, subgraph representations reflect weather context within the $k$-hop radius of NWP points. 
Finally, the $\textrm{MLP}(\cdot)$ layer maps high dimensional subgraph representations to the weather variables and predicts the current conditions. 

The objective of the fine-tuning is to accurately predict the current atmospheric states, aligning them as closely as possible with their actual values.
During training, the regression errors over all subgraphs and weather variables are computed using the L2 loss, which is defined as:
\begin{align}
	L_{reg} = \lVert Z, \hat{Z}\rVert_2^2+\psi\lVert W\rVert_2^2,
\end{align} 
where $Z$ and $\hat{Z}$ are the true and predicted atmospheric states, respectively.

\section{Observation Impact Analysis}

Estimating the impact of observations in meteorology involves understanding the contribution of a node \( v_j \) within a weather context \( g_m(v_i) \) to the predicted atmospheric states \( \hat{Z} \). 
The sensitivity of node \( v_j \) to the prediction \( \hat{Z}_{g_m(v_i)} \), represented as \( S_{i,j}(H^{(n)}_j,\hat{Z}_{g_m(v_i)}) \), is used to estimate this impact. 
The overall importance of observation \( v_j \) in the meteorological graph \( \mathcal{N} \) is then calculated by averaging sensitivities across different subgraphs:
\begin{align}
S_j = \frac{1}{i} \sum_{i}S_{i,j}.
\end{align} 
This approach allows us to aggregate the importance of each observation type, thereby determining the impact of each observation type on estimating atmospheric states.

To quantitatively measure the impact, we use gradients from the prediction model, commonly applied in graph reasoning.
Three methods are used to provide explanations based on the gradients and are empirically compared.  
The contrastive gradient-based saliency map (SA) \cite{Pope2019} utilizes these gradients to indicate how changes in the input could lead to variations in the output.
This can be formulated as:
\begin{align}
	L_{Gradient} = \textrm{ReLU} \left( \frac{\partial \hat{Z}}{\partial H^{(n)}} \right).
 \label{eq:L_grad}
\end{align} 

Additionally, the Grad-CAM method \cite{Pope2019} focuses on the last graph convolutional layers rather than the input space, identifying node importance using back-propagation gradients. This method computes weights \( \alpha_n \) as the average of the gradients and represents node importance as a weighted sum of feature maps:
\begin{align}
	L_{Grad-CAM} = \textrm{ReLU} \left( \sum_n \alpha_n H^{(n)} \right).
\end{align}

Layer-wise Relevance Propagation (LRP) \cite{baldassarre2019explainability} provides another perspective by reverse-propagating the prediction from the output of the graph convolutional layers back to the input features. It decomposes the prediction score into neuron importance scores based on hidden features and weights. The relevance score propagation at the neuron level is given by:
%
\begin{align}
	R_a = \sum_b \frac{h_a^{(n)} w_{ab}}{\sum_k h_k^{(n)} w_{kb}} R_b,
\end{align}
where \( h_a^{(n)} \) is the activation of the \( a^{th} \) hidden neuron in the \( n^{th} \) layer, and \( w_{ab} \) is the weight connecting the \( a^{th} \) neuron to the \( b^{th} \) neuron. This method intuitively assigns a larger fraction of the target neuron score to neurons contributing more significantly to the target neuron activation.

\section{Experimental Results and Discussion}
In this section, we validate the proposed atmospheric state estimation system and visualize the importance of observation.
We collected observation data and KIM data used in the Korea Meteorological Administration (KMA).
We used the output (i.e., u-component of wind (U), v-component of wind (V), temperature (T), relative humidity (Q)) of the DA system as the true value since the output was assumed to be the actual atmospheric state in the NWP system. 
Also, the observations were preprocessed by the KMA system. 
AIRCRAFT (U, V, T), GPSRO (banding angle (BA)), SONDE (U, V, T, Q), AMV (brightness temperature (TB)), AMSU-A (TB), AMSR2 (TB), ATMS (TB), CrIS (TB), GK2A (TB), IASI (TB), and MHS (TB), a total of 11 satellite and ground observations \cite{Kang2018} that have different variables (U, V, T, Q, TB, and BA) were used in the experiment. 
We have restricted the region to 500 hPa over East Asia.
The observation and NWP data (20 April 2021 to 30 April 2021) were used as the training dataset.
We then evaluated the proposed model using the next ten days (1 May 2021 to 10 May 2021).

\subsection{Effectiveness of the Proposed Estimation Model}

To explore the contribution of the graph-based data structure, we first compare the prediction performance of our model with fully connected networks (FCN).
In addition, we also evaluate the influence of the self-supervised learning and attention mechanisms through ablation tests. 

\begin{table}[h]
	\centering
	\footnotesize
	\begin{tabular}{c|l|ccccc}
		\toprule
		\textbf{Model}& \textbf{Variables}& \textbf{$RMSE$}& \textbf{$MAE$}& \textbf{$R^2$}& \textbf{$var$}\\
		\midrule
		\multirow{4}{*}{FCN}		
		& U(m/s)	& 0.20 &0.17&0.60&0.60\\
		& V(m/s)	& 0.13 &0.11&0.34&0.34\\
		& T(K)		& 0.23 &0.20&0.83&0.83\\
		& Q(kg/kg)	& 0.07 &0.07&0.37&0.37\\
		\midrule
		\multirow{4}{*}{GCN}		
		& U(m/s)	& 0.20 &0.17&0.64&0.64\\
		& V(m/s)	& 0.10 &0.07&0.56&0.56\\
		& T(K)		& 0.22 &0.20&0.88&0.88\\
		& Q(kg/kg)	& 0.05 &0.05&0.53&0.53\\
		\midrule
		\multirow{4}{*}{GAT}		
		& U(m/s)	& 0.18 &0.16&0.67&0.67\\
		& V(m/s)	& 0.10 &0.07&0.56&0.56\\
		& T(K)		& 0.22 &0.19&0.88&0.88\\
		& Q(kg/kg)	& 0.04 &0.04&0.56&0.56\\
		\midrule
		\multirow{4}{*}{\makecell{Proposed \\w/ GCN}}	
		& U(m/s)	& \textbf{0.16} &\textbf{0.13}&\textbf{0.73}&\textbf{0.74}\\
		& V(m/s)	& \textbf{0.09} &\textbf{0.06}&\textbf{0.73}&\textbf{0.73}\\
		& T(K)		& \textbf{0.19} &\textbf{0.16}&\textbf{0.93}&\textbf{0.93}\\
		& Q(kg/kg)	& \textbf{0.03} &\textbf{0.03}&\textbf{0.64}&\textbf{0.64}\\
		\midrule
		\multirow{4}{*}{\makecell{Proposed \\w/ GAT}}
		& U(m/s)	& 0.17 &0.14&0.72&0.72\\
		& V(m/s)	& 0.09 &0.07&0.71&0.72\\
		& T(K)		& 0.20 &0.18&0.90&0.90\\
		& Q(kg/kg)	& 0.04 &0.04&0.62&0.62\\
		\bottomrule
	\end{tabular}
\caption{A performance comparison of the proposed model with the baseline models. }
\label{table1}
\end{table}

In Table \ref{table1}, GCN, Graph Attention Network (GAT), and the proposed models have higher accuracy than the FCN model for all evaluation metrics which are widely used in the previous study \cite{Jeon2022}. 
Among the evaluation metrics, the accuracy of $var$ and $R^2$ show similar results, but they are calculated with different variances. 
Therefore, if the two values are the same, we can assess that the error of the model is unbiased.
We have verified that graph-structured meteorological data can improve the performance of current atmospheric state prediction. 
The proposed model based on self-supervised learning achieved significantly better performance than GCN and GAT.
A node representation that has been pre-trained on the node feature reconstruction considering meteorological contexts is more effective in estimating current atmospheric states than the cases without opportunities to learn spatial correlations between weather variables.
In particular, the performance of GAT, which considers nodes' importance based on their attention scores, is slightly better than GCN for all meteorological variables.
However, the U and T variables show very small performance deviations.
We assume that for estimating certain weather variables, all the closely located observations can have uniformly high (or low) importance.


On the other hand, GCN achieved better performance than GAT, as the backbone of our model. 
The proposed model based on GCN increased the accuracy by 14.06\%, 30.36\%, 5.68\%, and 20.75\% over vanilla GCN, while the proposed model based on GAT increased the accuracy by 7.46\%, 26.79\%, 2.27\%, and 10.71\% over vanilla GAT for each variable, respectively.
Pre-training the feature reconstruction task enables GNN models to understand correlations between weather variables. 
Then, the accuracy of GCN could be significantly improved, since GCN models merely aggregate neighboring nodes' features with mean aggregator.

\subsection{Stability of the Explainability Methods}

The results of the explainability methods provide insights into correlations between weather variables in a human-understandable way. 
However, it is difficult to evaluate these methods from this perspective due to the lack of ground truth and criteria for human understandability.
Also, comparing different explainability methods requires a lot of time and human resources to investigate the results for each meteorological context subgraph.
Therefore, evaluation metrics should evaluate the results from the perspective of the model, such as whether the explanations are faithful to the model.
We define $Fidelity+$ as the difference in accuracy obtained by occluding fixed percentages of input features assessed as important.
In addition, $Fidelity-$ indicates the difference between the predictions obtained by masking unimportant input features while retaining the important features.
We applied the explanation method to the proposed model based on GCN, which has the best performance in atmospheric state estimation.

\begin{table}[h]
	\centering
	\footnotesize
	\begin{tabular}{c|cc|cc}
		\toprule
        \multirow{2}{*}{Methods}& \multicolumn{2}{c|}{10\%} & \multicolumn{2}{c}{20\%}\\
		&  {$Fidelity+$}& $Fidelity-$&{$Fidelity+$}& $Fidelity-$\\
		\midrule
        SA         &   0.17    & 0.09  &   0.35    & 0.19  \\
		Grad-CAM   &   0.20    & \textbf{0.08}  &   0.36    & 0.17  \\
		LRP&   \textbf{0.25}    & 0.09      &   \textbf{0.39}    & \textbf{0.16}  \\
		\bottomrule
	\end{tabular}
\caption{Fidelity comparisons among explainability methods.}
\label{table2}
\end{table}

In Table~\ref{table2}, we occlude $10\%$ and $20\%$ of the input features and then compare their fidelity scores.
In the $Fidelity+$ metric, LRP outperforms the other explainability methods, showing that the propagation-based technique is more desirable than traditional gradient-based techniques for GNNs.
We can assume that SA and Grad-CAM have not made a sophisticated transition from the image domain to the graph-specific design, which is the main reason for their lower performance than the propagation-based LRP methods.
The $Fidelity-$ scores are similar to the $Fidelity+$ results, but the performance differences between the models are small.
For low importance nodes, all methods perform consistently in estimating the node importance.
When occluding nodes in the bottom 20\% of importance, the $Fidelity+$ score is significantly large due to the loss of a large number of nodes. 
However, masking nodes in the bottom 10\% of importance does not have a significant impact on accuracy.

\begin{figure}[h]
	\centering
	\includegraphics[width=0.9\columnwidth]{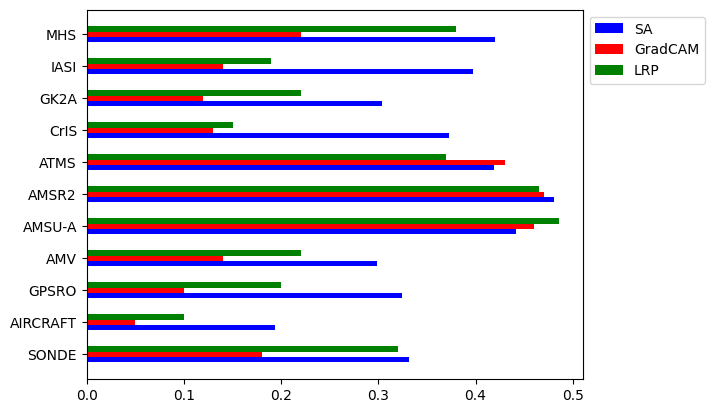} 
	\caption{The averaged impact of each observation type.}
	\label{fig1}
\end{figure}

As a result, the averaged effect of each type of observation on the estimation of the current state of the atmosphere can be visualized as shown in Figure~\ref{fig1}.
The variation of importance by the observation type assessed by SA is small compared to other methods.
Node importance is widely distributed across observation types because the SA method has explainable noise and poor localization performance \cite{Pope2019}.
The Grad-CAM, with its improved localization performance, has the largest variation in importance by observation type.
Therefore, the method has a high potential for use in cases where $n$ observation types with very large impacts or $n$ observation types with very small impacts need to be selected. 
Although the LRP method estimates different degrees of impact for different observation types, the ranking of importance observation is similar.

\begin{figure}[h]
	\centering
      \subfigure[SA]{\includegraphics[width=0.9\columnwidth]{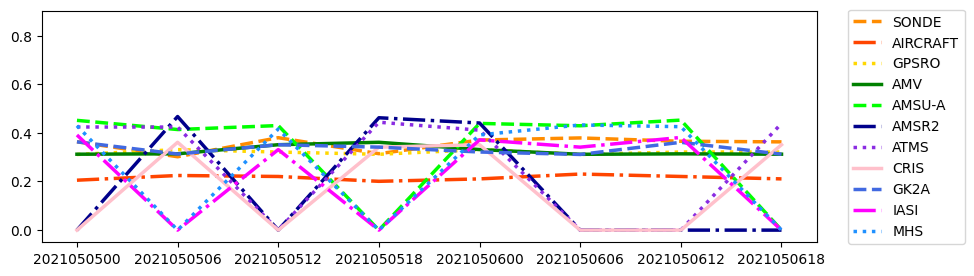}}
      \subfigure[Grad-CAM]{\includegraphics[width=0.9\columnwidth]{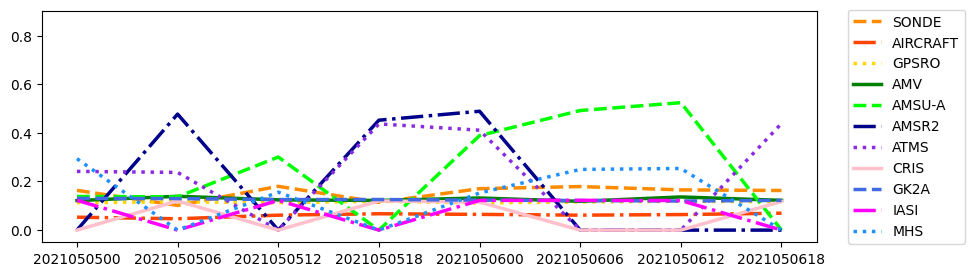}}
      \subfigure[LRP]{\includegraphics[width=0.9\columnwidth]{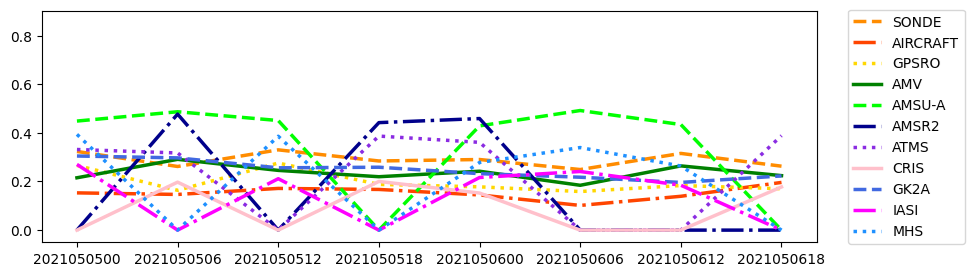}}
	\caption{Time series of observations impact.}
	\label{fig2}
\end{figure}

Figure~\ref{fig2} shows the impact by observation type evaluated for each time period, and observation types with an importance of 0 are those for which there are no observations exist in the target region (i.e., East Asia) at that time. 
The time series pattern of the impact by observation type evaluated by SA and Grad-CAM in Figure~\ref{fig2} does not change significantly. 
In addition, compared to the SA method, the Grad-CAM method assigns relatively greater importance to high-impact observation types, and the high-impact observation types are emphasized.
The LRP method calculates the relative importance for each observation type, and the impact of each observation is evaluated differently for each meteorological context.
The property of conservation in the LRP method is significant in interpreting the physical attributions in the graph-level prediction task.

\section{Conclusion}
In this paper, we propose an atmospheric state estimation model based on self-supervised graph neural networks. 
Then, we applied explainability methods to analyze the impact of observations on the current atmospheric state and visualize the importance of observation type.

\section*{Acknowledgments}

This work was supported 
in part by the R\&D project “Development of a Next-Generation Data Assimilation System by the Korea Institute of Atmospheric Prediction System (KIAPS)”, funded by the Korea Meteorological Administration (KMA2020-02211) (H.-J.J.)
and in part by the National Research Foundation of Korea (NRF) Grant funded by the Korea government (MSIT) (No. 2022R1F1A1065516 and No. 2022K1A3A1A79089461) (O.-J.L.).

\bibliographystyle{unsrt} 
\bibliography{aaai24}

\begin{thebibliography}{10}

\bibitem{Stulec2019}
Ivana Štulec, Kristina Petljak, and Dora Naletina.
\newblock Weather impact on retail sales: How can weather derivatives help with adverse weather deviations?
\newblock {\em Journal of Retailing and Consumer Services}, 49:1--10, July 2019.

\bibitem{Kotsuki2019}
Shunji Kotsuki, Kenta Kurosawa, and Takemasa Miyoshi.
\newblock On the properties of ensemble forecast sensitivity to observations.
\newblock {\em Quarterly Journal of the Royal Meteorological Society}, 145(722):1897--1914, May 2019.

\bibitem{Kwon2018}
In-Hyuk Kwon, Hyo-Jong Song, Ji-Hyun Ha, Hyoung-Wook Chun, Jeon-Ho Kang, Sihye Lee, Sujeong Lim, Youngsoon Jo, Hyun-Jun Han, Hanbyeol Jeong, Hui-Nae Kwon, Seoleun Shin, and Tae-Hun Kim.
\newblock Development of an operational hybrid data assimilation system at kiaps.
\newblock {\em Asia-Pacific Journal of Atmospheric Sciences}, 54(S1):319--335, June 2018.

\bibitem{Kang2018}
Jeon-Ho Kang, Hyoung-Wook Chun, Sihye Lee, Ji-Hyun Ha, Hyo-Jong Song, In-Hyuk Kwon, Hyun-Jun Han, Hanbyeol Jeong, Hui-Nae Kwon, and Tae-Hun Kim.
\newblock Development of an observation processing package for data assimilation in kiaps.
\newblock {\em Asia-Pacific Journal of Atmospheric Sciences}, 54(S1):303--318, June 2018.

\bibitem{Kalnay2012}
Eugenia Kalnay, Yoichiro Ota, Takemasa Miyoshi, and Junjie Liu.
\newblock A simpler formulation of forecast sensitivity to observations: application to ensemble kalman filters.
\newblock {\em Tellus A: Dynamic Meteorology and Oceanography}, 64(1):18462, December 2012.

\bibitem{Buehner2018}
Mark Buehner, Ping Du, and Joël Bédard.
\newblock A new approach for estimating the observation impact in ensemble–variational data assimilation.
\newblock {\em Monthly Weather Review}, 146(2):447--465, January 2018.

\bibitem{Jeon2021}
Hyeon-Ju Jeon and Jason~J Jung.
\newblock Discovering the role model of authors by embedding research history.
\newblock {\em Journal of Information Science}, 49(4):990--1006, August 2021.

\bibitem{Hoang2023}
Van~Thuy Hoang, Hyeon-Ju Jeon, Eun-Soon You, Yoewon Yoon, Sungyeop Jung, and O-Joun Lee.
\newblock Graph representation learning and its applications: A survey.
\newblock {\em Sensors}, 23(8):4168, April 2023.

\bibitem{Lee2021}
O-Joun Lee, Hyeon-Ju Jeon, and Jason~J. Jung.
\newblock Learning multi-resolution representations of research patterns in bibliographic networks.
\newblock {\em Journal of Informetrics}, 15(1):101126, February 2021.

\bibitem{Jeon2022}
Hyeon-Ju Jeon, Min-Woo Choi, and O-Joun Lee.
\newblock Day-ahead hourly solar irradiance forecasting based on multi-attributed spatio-temporal graph convolutional network.
\newblock {\em Sensors}, 22(19):7179, sep 2022.

\bibitem{Ma2023}
Minbo Ma, Peng Xie, Fei Teng, Bin Wang, Shenggong Ji, Junbo Zhang, and Tianrui Li.
\newblock {HiSTGNN}: Hierarchical spatio-temporal graph neural network for weather forecasting.
\newblock {\em Information Sciences}, 648:119580, nov 2023.

\bibitem{Yang2018}
Yuting Yang, Junyu Dong, Xin Sun, Estanislau Lima, Quanquan Mu, and Xinhua Wang.
\newblock A cfcc-lstm model for sea surface temperature prediction.
\newblock {\em IEEE Geoscience and Remote Sensing Letters}, 15(2):207--211, February 2018.

\bibitem{Lam2023}
Remi Lam, Alvaro Sanchez-Gonzalez, Matthew Willson, Peter Wirnsberger, Meire Fortunato, Ferran Alet, Suman Ravuri, Timo Ewalds, Zach Eaton-Rosen, Weihua Hu, Alexander Merose, Stephan Hoyer, George Holland, Oriol Vinyals, Jacklynn Stott, Alexander Pritzel, Shakir Mohamed, and Peter Battaglia.
\newblock Learning skillful medium-range global weather forecasting.
\newblock {\em Science}, page eadi2336, November 2023.

\bibitem{Yuan2022}
Hao Yuan, Haiyang Yu, Shurui Gui, and Shuiwang Ji.
\newblock Explainability in graph neural networks: A taxonomic survey.
\newblock {\em IEEE Transactions on Pattern Analysis and Machine Intelligence}, 45(5):5782--5799, 2023.

\bibitem{Pope2019}
Phillip~E. Pope, Soheil Kolouri, Mohammad Rostami, Charles~E. Martin, and Heiko Hoffmann.
\newblock Explainability methods for graph convolutional neural networks.
\newblock In {\em Proceedings of the 2019 {IEEE}/{CVF} Conference on Computer Vision and Pattern Recognition ({CVPR 2019})}. {IEEE}, jun 2019.

\bibitem{Ying2019}
Zhitao Ying, Dylan Bourgeois, Jiaxuan You, Marinka Zitnik, and Jure Leskovec.
\newblock Gnnexplainer: Generating explanations for graph neural networks.
\newblock In {\em Proceedings of the 32th Advances in Neural Information Processing Systems (NeurIPS 2019), 8-14 December 2019, Vancouver, BC, Canada}, pages 9240--9251, 2019.

\bibitem{Vu2020}
Minh~N. Vu and My~T. Thai.
\newblock Pgm-explainer: Probabilistic graphical model explanations for graph neural networks.
\newblock In {\em Proceedings of the 33th Advances in Neural Information Processing Systems (NeurIPS 2020), 6-12 December 2020, virtual}, 2020.

\bibitem{Irvine2011}
E.~A. Irvine, S.~L. Gray, J.~Methven, and I.~A. Renfrew.
\newblock Forecast impact of targeted observations: Sensitivity to observation error and proximity to steep orography.
\newblock {\em Monthly Weather Review}, 139(1):69--78, January 2011.

\bibitem{baldassarre2019explainability}
Federico Baldassarre and Hossein Azizpour.
\newblock Explainability techniques for graph convolutional networks.
\newblock arXiv preprint arXiv:1905.13686, 2019.

\end{thebibliography}

\end{document}